\pdfoutput=1

\documentclass[11pt]{article}
\usepackage[]{emnlp2023}
\usepackage{adjustbox}
\usepackage{todonotes}
\usepackage{times}
\usepackage{latexsym}
\usepackage[T1]{fontenc}
\usepackage[utf8]{inputenc}
\usepackage{microtype}
\usepackage{inconsolata}
\usepackage{multirow}
\usepackage{xcolor}
\usepackage{graphicx}
\usepackage{caption}
\usepackage{subcaption}
\usepackage{booktabs}
\usepackage{multirow}
\usepackage{listings}

\makeatletter
\newcommand\subsubsubsection{\@startsection{paragraph}{4}{\z@}%
  {-3.25ex\@plus -1ex \@minus -.2ex}%
  {1.5ex \@plus .2ex}%
  {\normalfont\normalsize\bfseries}}
\makeatother

\title{Detrimental Contexts in Open-Domain Question Answering}

\author{Philhoon Oh\\KAIST AI\\\url{philhoonoh@kaist.ac.kr} \And James Thorne\\KAIST AI\\\url{thorne@kaist.ac.kr}}

\begin{document}
\maketitle

\begin{abstract}
For knowledge intensive NLP tasks, it has been widely accepted that accessing more information is a contributing factor to improvements in the model's end-to-end performance. However, counter-intuitively, too much context can have a negative impact on the model when evaluated on common question answering (QA) datasets.  In this paper, we analyze how passages can have a detrimental effect on retrieve-then-read architectures used in question answering. Our empirical evidence indicates that the current read architecture does not fully leverage the retrieved passages and significantly degrades its performance when using the whole passages compared to utilizing subsets of them.  
Our findings demonstrate that model accuracy can be improved by 10\% on two popular QA datasets by filtering out detrimental passages. Additionally, these outcomes are attained by utilizing existing retrieval methods without further training or data.
We further highlight the challenges associated with identifying the detrimental passages. First, even with the correct context, the model can make an incorrect prediction, posing a challenge in determining which passages are most influential.
Second, evaluation typically considers lexical matching, which is not robust to variations of correct answers.
Despite these limitations, our experimental results underscore the pivotal role of identifying and removing these detrimental passages for the context-efficient retrieve-then-read pipeline. \footnote{Code and data are available on \url{https://github.com/xfactlab/emnlp2023-damaging-retrieval}}

\end{abstract}

\section{Introduction}
Knowledge-intensive NLP tasks such as open-domain question answering \citep{rajpurkar-etal-2016-squad, yang-etal-2018-hotpotqa} and evidence-based fact verification \citep{thorne-etal-2018-fever, jiang-etal-2020-hover} require models to use external sources of textual information to condition answer generation in response to an input. This family of tasks shares a common architecture for modelling \citep{NEURIPS2020_6b493230, petroni-etal-2021-kilt, thakur2021beir, izacard2022few, lyu2023improving}  where a two-stage architecture of a \textit{retriever} model finds contextual passages and passes these to a \textit{reader} module that uses this context when it generates an answer.  

In this two-step pipeline, expected performance (such as answer \textit{exact match} score) increases monotonically as more information is provided to the reader model at test time. Therefore, it is commonly conceived that providing the reader with more context improves overall performance \citep{izacard2020leveraging, liu2023lost}. However, empirical studies \citep{10.1145/3477495.3532034} contradict this general belief:  even when true positive passages are provided to the model, the presence of relevant but non-evidential documents can change correct outputs to incorrect ones in a phenomena called \textit{damaging retrieval}. While this phenomena was not thoroughly explored in the context of question answering, it could suggest that the current reader may be influenced by \textit{damaging retrieval}, potentially leading to a decline in performance.

\begin{figure}[!t]
  \centering
  \includegraphics[width=0.8\linewidth]{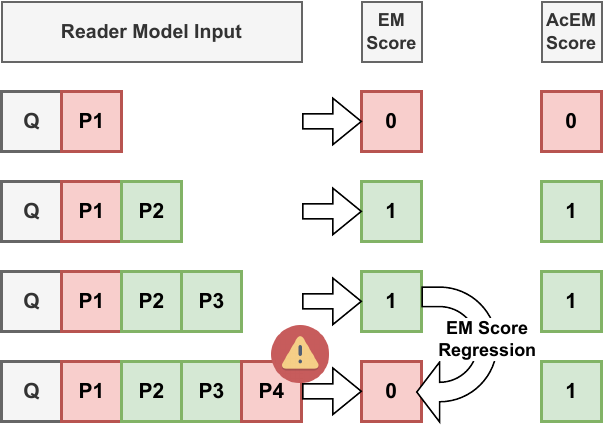}
  \caption{Even when provided with correct passage (P2) addition of further information (P4) causes the reader model to change the prediction, highlighting issues when using recall-optimized retrievers. Differences between Exact Match (EM) score and Accumulated EM (AcEM) indicate knowledge issues.}
  \label{fig:intro}
\end{figure}

In this paper, we propose an approach to addressing this problem by identifying and removing contexts that are detrimental to the performance of the model. As described in Figure \ref{fig:intro}, we treat the reader model, FiD \citep{izacard2020leveraging}, as a black-box oracle and identify which passages are damaging by evaluating how different subsets of the retrieved context cause the answer to change.  Our experiments are conducted on two question answering tasks: Natural Questions \citep{kwiatkowski-etal-2019-natural} and TriviaQA \citep{joshi-etal-2017-triviaqa}, evaluating the effect of three IR systems: DPR \citep{karpukhin-etal-2020-dense},  SEAL \citep{bevilacqua2022autoregressive}, and Contriever \citep{izacard2021contriever}. 

Our findings challenge the conventional assumption that more passages lead to higher performance. In fact, introducing additional relevant contexts can actually worsen the performance of the model in question answering. By excluding passages that have a detrimental effect, we observe up to 10\% improvement in the exact match score under ideal conditions without requiring any architectural modifications. Notably, these gains surpass the performance improvements seen in recent works that propose FiD extensions like FiD-light\citep{hofstatter2022fid} and FiE \citep{https://doi.org/10.48550/arxiv.2211.10147}. In addition, we analyze why automatically identifying damaging contexts is challenging, even with attention-based and model-based proxies for filtering. Despite these inherent limitations, considering effectiveness and operational efficiency, our empirical findings could potentially direct the focus of open-domain NLP models and yield performance improvements when using fewer context passages.

\section{Background}

\subsection{Knowledge Intensive NLP}
To integrate external knowledge for NLP tasks such as question answering \citep{kwiatkowski-etal-2019-natural, joshi-etal-2017-triviaqa, yang-etal-2018-hotpotqa}, slot filling \citep{levy-etal-2017-zero}, and fact verification \citep{thorne-etal-2018-fever}, systems employ a retrieve-then-read pipeline of two models \citep{chen-etal-2017-reading}. A retriever first selects the context passages over a large corpus such as Wikipedia and then feeds information to the reader to derive the answer. Consequently, improving the recall of the retriever leads to improvement of the downstream reader.

Large-scale information fusion architectures such as Fusion in Decoder (FiD) \cite{izacard2020leveraging} are able to combine information in multiple passages when decoding the answer. In contrast to the conventional T5-Model \cite{DBLP:journals/corr/abs-1910-10683}, each (query, passage) pair is independently encoded, and the encoded vectors are concatenated when input into the decoder model. This allows more documents to be encoded overcoming the high memory complexity of self-attention in the encoder. Experimental results indicate that this architecture also supports recall-oriented optimization of the upstream information retrieval systems. For question-answering tasks, the answer exact match scores increased monotonically with the number of context passages used \citep{izacard2020leveraging, liu2023lost}.

Extensions of this architecture, such as FiD-Light \cite{hofstatter2022fid}, FiDO \cite{https://doi.org/10.48550/arxiv.2212.08153} are introduced to further increase the runtime efficiency. FiD-Light, for example, compresses passage encoder embedding into first-K vectors to overcome a performance bottleneck in the decoder. In the case of FiDO, it only keeps cross-attention on every K-th decoder layer as well as applies multi-query attention to mitigate the inference costs and memory usage. Unlike these architectures, FiE \cite{https://doi.org/10.48550/arxiv.2211.10147}, an encoder-only model, adds global encoder layers to fuse the information across multiple evidence passages. While these models significantly reduce inference time, they only result in minor increases in the answer exact match score.

Recent advances in LLMs demonstrate remarkable capabilities in natural language generation tasks \cite{liang2022holistic}. Within the context of question answering, \citealt{lyu2023improving} utilize an LLM as a reader in a retrieval-augmented generation manner \cite{NEURIPS2020_6b493230}. However, recent studies \citep{zheng2023large} demonstrate that answers generated from LLMs vary depending on the order of given contexts/answer choices and are sometimes lost in the middle \citep{liu2023lost}. While \citealt{zheng2023large} proposes a method for alleviating this order-variant issue by permutating given choices, this approach is not practical for 100 candidate contexts. Although investigating this issue falls outside the scope of our research, further research is needed to address the issue of the answer variability in LLMs, which arises from the order of provided contexts/answer choices.

\subsection{Improving Retrieval}
Two different approaches have been studied for retrieving relevant contexts: one approach employs a bi-encoder architecture, where the inner product of query embeddings and context embeddings is used as a proxy for the relevance score. ORQA\cite{DBLP:journals/corr/abs-1906-00300}, DPR \citep{karpukhin-etal-2020-dense} are both based on BERT-based bi-encoder architectures \citep{DBLP:journals/corr/abs-1810-04805}. Contriever~\citep{izacard2021contriever} trains a single encoder in an unsupervised manner to represent both query and context embeddings. These encodings capture the relation between query and passages where most similar results can be identified through ranking the top-$N$ articles through inner-product search between embedded passages and the embedding of the query.  

Alternatively, generative retrieval offers an alternative mechanism where the relationship between the query and an entity is encoded in the parametric space within a model. GENRE \citep{DBLP:journals/corr/abs-2010-00904} utilizes an autoregressive sequence-to-sequence model to generate canonical entity titles for a query with constrained decoding using a prefix tree. Extending this, SEAL \citep{bevilacqua2022autoregressive} directly predicts substrings in the corpus using an FM-index.  GenRead \cite{yu2023generate} does not rely on any indexing system: LLMs are used to generate synthetic contexts. These approaches demonstrate favorable precision-recall trade-offs while requiring smaller index storage footprints.

\subsection{Damaging Retrieval}
\label{sec:damagingRetrieval}
Conventionally, retrieval systems are optimized to find maximally relevant documents to support a given user query \citep{ferrante2018general}. However, \citet{10.1145/1099554.1099646} demonstrates that the retriever can extract relevant but harmful documents and \citet{10.1145/3477495.3532034} also show that irrelevant or incorrect documents added to the input of the model can negatively affect its performance. 

Identifying these damaging passages in a two-stage pipeline has not been extensively explored. However, one can consider reranking \citep{iyer2020reconsider, kongyoung-etal-2022-monoqa, glass-etal-2022-re2g} as a process for selecting an optimal subset of retrieved passages and removing detrimental ones. However, FiD architectures scale well with many passages, enabling higher exact match scores simply by increasing the number of passages provided to the model. Given a large enough budget for passages, re-ranking this set of passages may not necessarily exclude them from the answer set. Furthermore, as further discussed in Section~\ref{label:dmgpsg2}, the FiD model is order invariant and the ranking of passages is ignored during encoding. 
In architectures such as FiD that treat retrieved evidence as a set, filtering the damaging passages is more appropriate than re-ranking the retrieval results.

\section{Motivating Pilot Study}
\label{label:dmgpsg1}
We demonstrate the effect of damaging passages extending \citet{10.1145/3477495.3532034} for the FiD model demonstrating that the model performance is sensitive to irrelevant context information. To select damaging context passages, we employ two methods: \emph{random sampling} and \emph{negative sampling}. For random sampling, we select passages from the corpus, $\mathcal{C}$, uniformly at random, while for negative sampling, we use BM25 to select passages that do not contain the correct answer but have high lexical overlaps. Results are reported in Figure~\ref{fig:damagingexpl} an Table~\ref{tab:ser}. 
We use a context of up to 5 passages on 2539 instances from the NQ dev dataset and employ FiD-large\footnote{\url{https://github.com/facebookresearch/fid}} trained on NQ train set.  
We add between 0-4 sampled passages in addition to the gold passage from the dataset.
FiD is invariant to passage ordering: we derive this analytically from lack of positional encoding between the encoded passages and validate it empirically through random permutation of the context set.

\subsection{Effects of Simulated Damaging Passages}
\label{label:dmgpsg2}
We compute Stability Error Ratio \citep[SER]{DBLP:journals/corr/abs-2108-11357}: the ratio of instances with modified outputs to additional samples, given the instance has correct predictions using one gold passage.

\begin{figure}[t]
\centering
    \begin{subfigure}[h]{0.49\linewidth}
        \includegraphics[width=\linewidth]{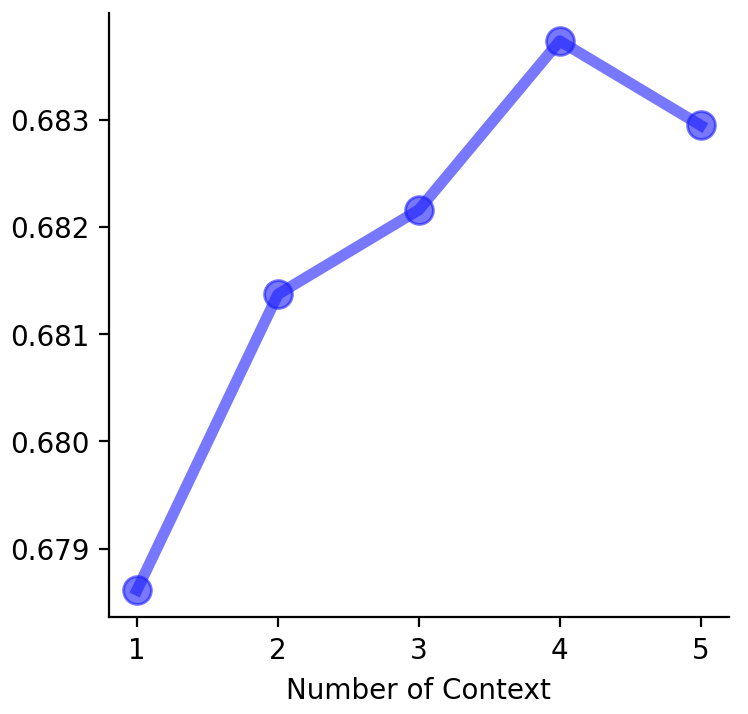}
        \caption{Random passages}
        \label{fig:fig1}
    \end{subfigure}
        \begin{subfigure}[h]{0.49\linewidth}
        \includegraphics[width=\linewidth]{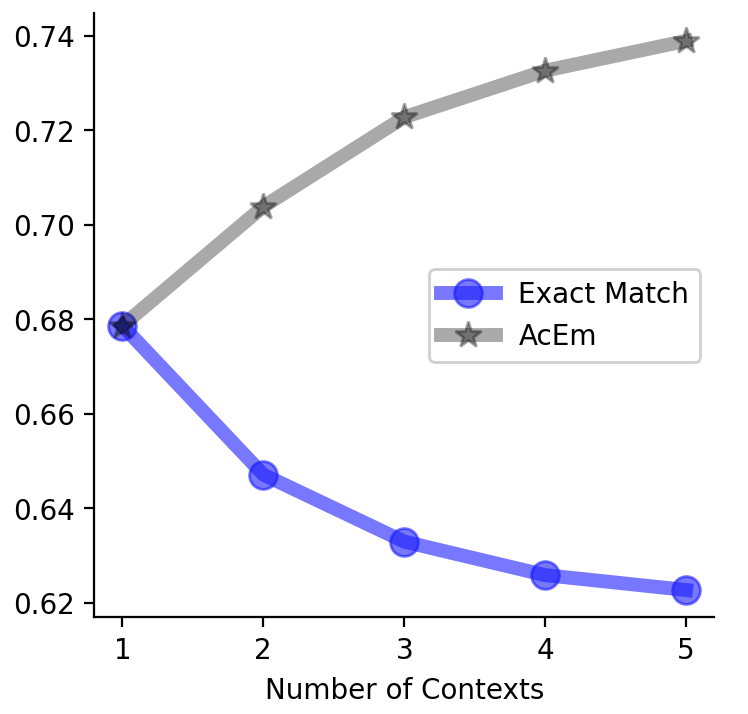}
        \caption{BM25 negative passages}
        \label{fig:fig2}
    \end{subfigure}
\caption{Exact Match scores of FiD on Natural Questions with a different number of contexts (x-axis). As the Exact Match score remains unchanged regardless of the position of the gold context, we represent this relationship using a single blue line in the plot.}
\label{fig:damagingexpl}
\end{figure}

\paragraph{Random Sampling}
Random passages should not contain information related to the query.. Where models are resilient to noise, we expect that randomly sampled additional do not drastically affect the reader model. Because of the large semantic distance between the randomly sampled texts and the gold passage, irrelevant information is not damaging. Injecting these random passages had negligable increase on answer EM (Figure~\ref{fig:damagingexpl}(a)) and SER is under 3$\%$ across all the cases in Table~\ref{tab:ser}. This, in fact, indicates that FiD is robust to additional random passages. 

\paragraph{Negative Sampling}
Passages sampled by BM25 have high lexical overlap and are semantically similar meaning to the query. We expected this to result in confusion in the model as the model uses related but insufficient information to condition answer generation. Adding additional negative samples results in a monotonic decrease in EM score, which is represented as a blue line in Figure~\ref{fig:damagingexpl}(b) and an increase in SER, reported in Table~\ref{tab:ser}. A rising number of instances where the model changes the answer from predicting the correct answer to an incorrect highlights the damaging effects.

In general, negative passages tend to have a negative impact on the model's inference. However, we have discovered that certain negative passages can actually enhance the model's performance when paired with gold evidence. To measure this, we calculate an extension of EM score called Accumulated Exact Match(AcEM). This checks the existence of Exact Match on up to top-K passages. Surprisingly, AcEM monotonically improves with additional negative passages, yielding higher results than when using only gold contexts. This demonstrates that the model can also benefit from negative samples to make correct predictions and implies that \textit{damaging} passages should be differentiated from \textit{negative} contexts. We report comparisons of AcEM and EM with respect to the number of context sizes in Figure~\ref{fig:damagingexpl}(b). 

\begin{table}[t]
\small
\centering
\begin{adjustbox}{scale=0.9}
\begin{tabular}{|c|c|c|c|c|}
\hline
\textbf{\# Negative Samples} & \textbf{1}      & \textbf{2}      & \textbf{3}      & \textbf{4}      \\ \hline
SER w/ random       & 0.75\% & 1.4\%  & 1.7\%  & 2.0\%  \\ \hline
SER w/ BM25         & 8.3\%  & 12.5\% & 14.8\% & 15.9\% \\ \hline
\end{tabular}
\end{adjustbox}
\caption{Stability Error Ratio on Random/BM25 negative passages. An increase in SER implies the model's instability on negative samples. 
}
\label{tab:ser}
\end{table}

\section{Damaging Passages from Retrievers}
\label{label:dmgpsR}
Section~\ref{label:dmgpsg1} simulates the impact of damaging passages. However, these negative passages are sampled from a distribution different from what the reader is trained on. Therefore, we experiment with retrieval models to confirm the presence of damaging effects in the retrieve-then-read pipeline.

\begin{figure*}[t]
  \centering
  \includegraphics[width=0.75\linewidth]{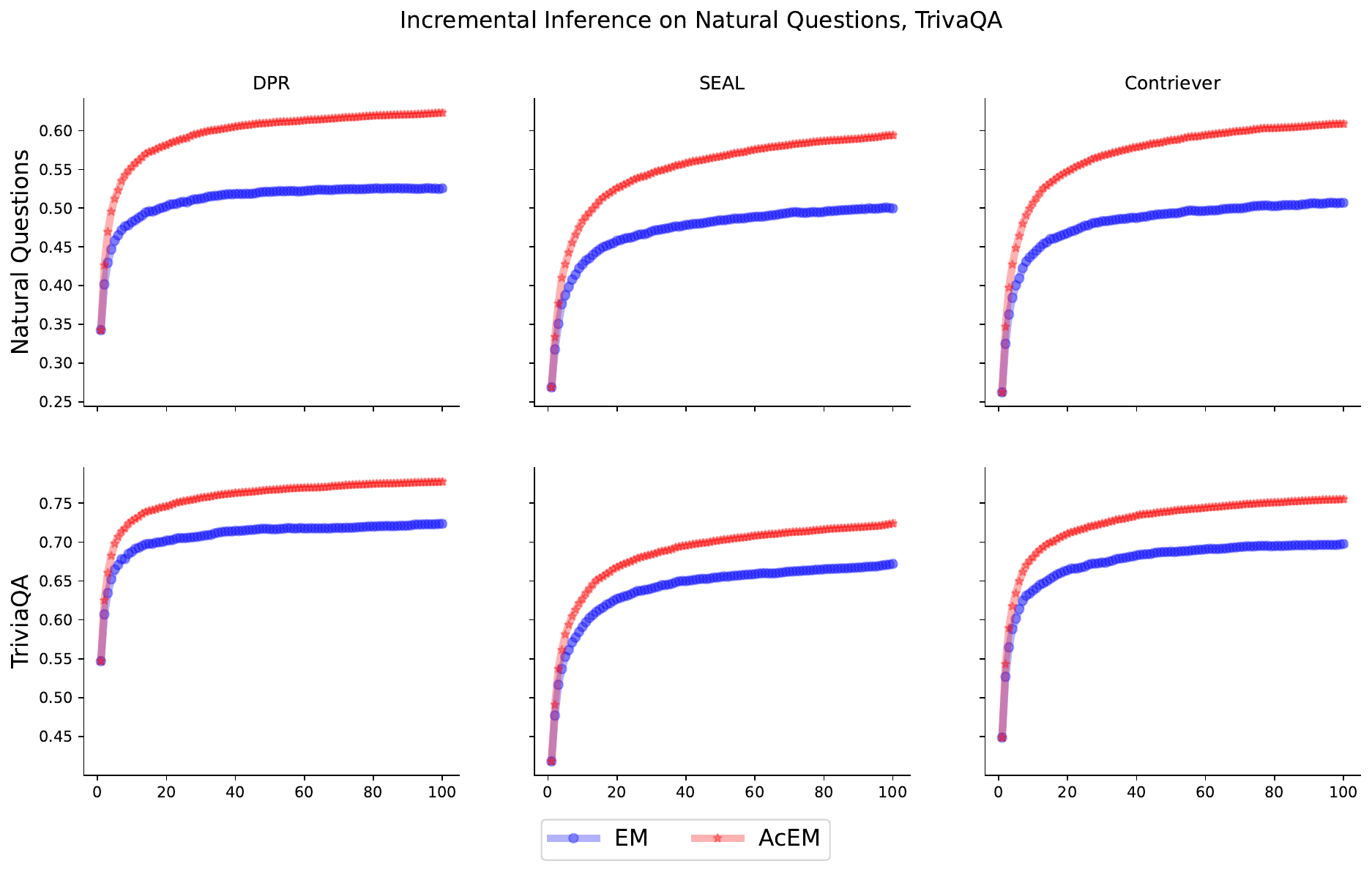}
  \caption{Incremental inference for 3 retrievers on NQ and TriviaQA dev set. This implies the existence of damaging passages in retrieved contexts, and higher performance can be attained using a subset of context.}
  \label{fig:incrementalinferenceplot1}
\end{figure*}

While retrievers predict a set of passages that are relevant to the model, it does not have perfect precision, and we do not have knowledge of which passages from the retriever are beneficial or detrimental. In order to identify wheter a passage is damaging or not, we utilize the reader model as a black box oracle to distinguish damaging contexts from positive contexts. Starting with the top-1 passage, we incrementally add passages from the retrieved list and evaluate how the answer changes. If the inference is correct with the top-1, it is classified as a positive context. We then proceed to evaluate the top-1 and top-2 passages together. If those passages generate an incorrect answer, the second evidence passage will be considered damaging. This is possible because of the order invariance property of FiD. Therefore, the process can be repeated iteratively for up to the top-N passages. We then use this sequence of predictions to yield the \textit{exact match pattern}. The exact match pattern is a binary sequence where a 1 represents that the k-th prediction matches the answer, and 0 otherwise (Figure~\ref{fig:intro}). This string represents an exact match at k (EM@k) when using up to top-k passages as input. From this, we also compute accumulated exact match (AcEM). AcEM@k is the maximum of all EM@k values up to $k$. For instance, AcEM@k equals 1 if at least one of the values in (EM@1, EM@2, ..., EM@k) equals 1. Thereby, the discrepancy between AcEM@k and EM@k can be used to measure the damaging effects on retrievers.

We utilize three different retrievers for two question answering validation datasets, Natural Questions \citep[NQ]{kwiatkowski-etal-2019-natural} and TriviaQA \citep[TQA]{joshi-etal-2017-triviaqa}, which were evaluated on FiD \cite{izacard2020leveraging}. Following previous work, we use a corpus of non-overlapping 100 word chunks for retrieval. We use the top-100 candidate passages retrieved by DPR\footnote{\url{https://github.com/facebookresearch/DPR}} \citep{karpukhin-etal-2020-dense}, SEAL\footnote{\url{https://github.com/facebookresearch/SEAL}} \citep{bevilacqua2022autoregressive}, and Contriever \citep{izacard2021contriever}\footnote{\url{https://github.com/facebookresearch/contriever}}. For DPR, we utilize the published retrieval result, while for the other retrievers, we run the code using the default parameters. We use two published models of FiD-large that were trained on DPR retrieved contexts from NQ/TriviaQA for inference. 

Results are illustrated in Figure~\ref{fig:incrementalinferenceplot1}. All models show higher AcEM@k on all k values. For NQ, the EM@100 scores from DPR, SEAL, and Contriever are 52.5\%, 50.0\%, and 50.7\% while attaining 62.3\%, 59.4\% and 60.8\% in AcEM@100. This indicates the existence of damaging effects in retrieved contexts, and the model performance could be at least 9\% points higher if the retrieved passages can be refined to remove damaging passages.

\section{Analysis Methodology}
\label{method}

\subsection{Identifying Passage Types}
\label{sec:probing}
In order to identify which passages are responsible for damaging effects in the retrieved list, we categorize passages based on the Exact Match(EM) pattern as an indicator to determine if a passage is positive or not. If the EM@k is equal to 1, it can be inferred that the kth passage is positive. However, this does not necessarily imply that the passage contains relevant information for the outcome. It could be a false positive due to the presence of preceding positive passages in the input. Similarly, a 0 in the EM pattern may suggest that the passage is negative. However, this could also be a false negative resulting from FiD's failure to predict the answer despite the presence of a correct passage.

What can be inferred from the EM pattern is when there is a change in value. If there is a transition from 0 to 1 in the EM pattern, we can infer that the passage corresponding to 1 contains at least some positive information for the outcome. Conversely, a transition from 1 to 0 suggests that the additional passage negatively impacts the prediction. Thus, we classify passages into five types based on transitions in the EM patterns: IZ(initial zeros), DP(definite positives), DN(definite negatives), SP(semi-positive), and SN(semi-negative). Let EM($p_{k}$) denote EM@k. Then, for each $p_{i}$ given $\mathbf{P} = \{p_{1}, p_{2}, ... , p_{N} \}$, we define  $\mathbf{Type}(p_{i})$:

\begin{itemize}
    \item \textbf{IZ} 
where EM($p_{i}$) = 0 and EM($p_{j}$) = 0 for all $j$ = $1,2,..,i-1$. Consecutive zeros appearing at the start of the EM pattern correspond to possibly relevant or uninformative passages.
    \item \textbf{DP}  
where EM($p_{i}$) = 1 and EM($p_{i-1}$) = 0 $or$ EM($p_{i}$) = 1 when $i$ = 1. Definite positive passages cause EM to transition from 0 to 1 $or$ the first passage enabling the correct answer.
    \item \textbf{DN}   
where EM($p_{i}$) = 0 and EM($p_{i-1}$) = 1. Definite negative passages cause a transition from a correct prediction to an incorrect one.
    \item \textbf{SP} and \textbf{SN} are semi-positive/negative passages do not cause transition(EM($p_{i}$) = EM($p_{i-1}$)). Incremental inference with these passages does not reveal their utility. 
\end{itemize}

\subsection{Passage Type Selection}
\label{sec:psgtype}
\begin{figure}[!t]
  \centering
  \includegraphics[width=0.8\linewidth]{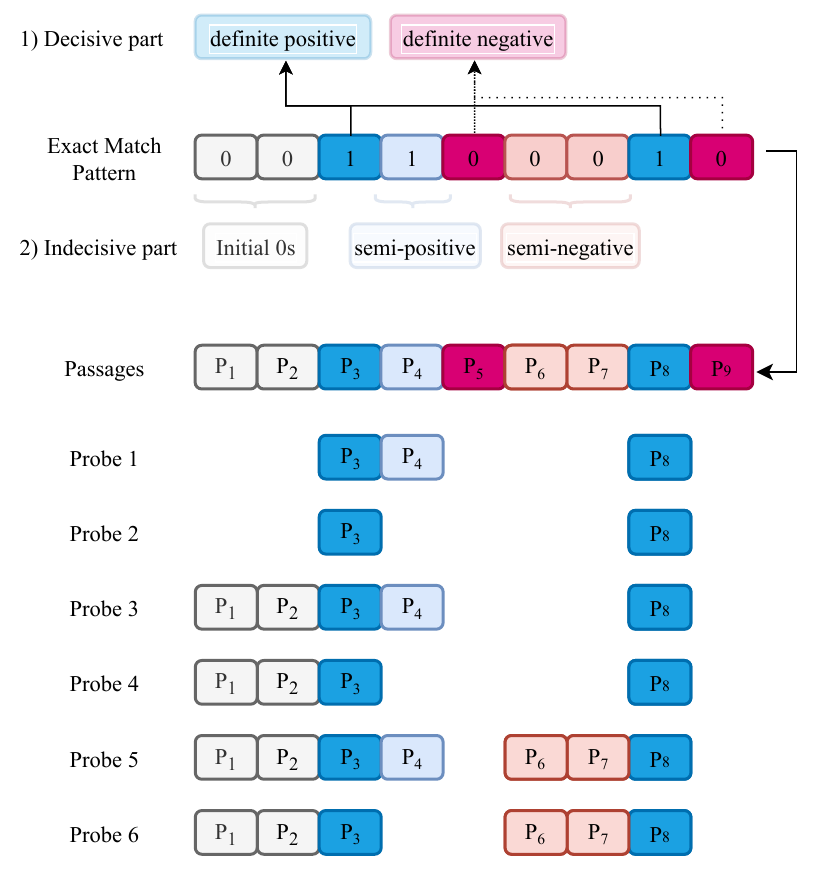}
  \caption{Context types and Probe Methods. Definite positive/negative occur on transitions of the EM pattern.}
  \label{fig:patterns}
\end{figure}

To assess the impacts of \textbf{DN} contexts along with different context types, we generate six probe patterns to feed into the model. These patterns consist of different combinations of the \textbf{DP}, \textbf{SP}, and \textbf{SN}. We also compare the effect of the \textbf{IZ} in our probe patterns too. In total, there are six probe patterns, illustrated in Figure~\ref{fig:patterns}. The most relevant approach to ours \citep{DBLP:journals/corr/abs-2112-08688} only performs leave-one-out generation: a passage is considered to be positive if the model fails to generate the correct inference without the passage. This approach aids identification of positive passages but does not consider which information can be detrimental to a model. Further elaboration on the distinctions with our methodology is provided in Appendix~\ref{appen:difference}.

\subsection{Model Analysis}
\label{sec:method-attention}
In Section~\ref{sec:psgtype}, we classify context types to assess their influence on the outcomes. In this section, we analyze \textbf{DP} and \textbf{DN} using two existing methods to check for distinct signals or patterns.

\paragraph{Attention Score Analysis}
Attention scores have been extensively studied to enhance the performance of both readers and retrievers. For instance, \citealt{xu2021attentionguided} employ cross-attention mechanisms to improve the performance of extractive readers. While \citealt{izacard2022distilling} use attention scores from the reader as a measure of passage contribution. Building on this insight, we compare the attention scores of \textbf{DP} and \textbf{DN} passages from the reader model.

\paragraph{Binary Classification}
Re-ranking is commonly used for enhancing the performance of the reader in the retrieve-then-read pipeline \citep{iyer2020reconsider, kongyoung-etal-2022-monoqa}. Additionally, \citealt{glass-etal-2022-re2g} demonstrates that a query-passage re-ranking (classification) can improve the final outcome. Inspired by this idea, we train three different encoders to train binary classification on \textbf{DP} and \textbf{DN} contexts. This enables us to check whether the models are capable of distinguishing between \textbf{DP} and \textbf{DN}.

\section{Experiments}
\label{sec:experiment}

\subsection{Probe-based Selection Inference}
We evaluate all 6 probing strategies from Section~\ref{sec:psgtype}. Initially we add padding passages and fix the context size to 100 during the inference. For probe 3, showing the highest EM score, we further experiment with varying context sizes of 5, 10, 20 and 40 on each DPR-retrieved test set. The remaining experimental setup follows Section~\ref{label:dmgpsR}.

\subsection{Attention Analysis}
Following the approach by \citealt{izacard2022distilling}, we calculate the cross-attention score by averaging across heads and layers of input tokens on the first output token. We conduct the experiment on top-20 DPR-retrieved contexts on NQ dev set. Next, we visualize the distribution of \textbf{DP}, \textbf{DN}, and other passage types using a density plot. We perform inference on the same dataset using passages with attention scores at various threshold values: 0.025, 0.05, 0.075, 0.1, and 0.2. We only consider contexts with an attention score higher than the threshold. If no context meets this criterion, we use the entire candidate passages instead.

\subsection{Binary Classification}
\label{exp:binary}
We train a binary classifier on the \textbf{DP} and \textbf{DN} contexts. To prepare the data, we split the DPR-retrieved NQ dev set into a 4:1 ratio, resulting in 7005 instances for training and 1752 instances for evaluation. From these instances, we extract 5205 \textbf{DP}s and 1516 \textbf{DN}s for training, while 1308 \textbf{DP}s and 396 \textbf{DN}s are used for evaluation. We make prediction using the question, title and context comparing a RoBERTa-large and T5-large encoder initialized with the huggingface repository checkpoints. We additionally compare using FiD-large encoder, which was pre-trained on DPR-retrieved NQ train datasets. For hyperparameters and settings, we use a batch size of 64, a learning rate of 5e-5, 3 epochs, and implement the fine-tuning process using the Huggingface transformer library\footnote{\url{https://huggingface.co/docs/transformers}}.

\section{Results and Analysis}

\subsection{Probe-based Selection Inference}
\label{sec:selection-inference}
\begin{table}[t]
\small
\centering
\begin{adjustbox}{scale=0.9}
\begin{tabular}{clcc|c}
\hline
\textbf{Dataset}               & \multicolumn{1}{l|}{\textbf{Retriever}}   & \textbf{EM@100} & \textbf{AcEM@100} & \textbf{Probe 3}  \\ \hline
                      & \multicolumn{1}{l|}{DPR}        & 52.5 & 62.3 & \underline{\textbf{61.8}}   \\
                      & \multicolumn{1}{l|}{SEAL}       & 50.0 & 59.4 & \textbf{52.9} \\
\multirow{-3}{*}{NQ}  & \multicolumn{1}{l|}{Contriever}  & 50.7 & 60.8 & \textbf{53.0} \\ \hline
                      & \multicolumn{1}{l|}{DPR}        & 72.3 & 77.7 & \underline{\textbf{77.6}}  \\
                      & \multicolumn{1}{l|}{SEAL}       & 67.1 & 72.3 & \textbf{72.5} \\
\multirow{-3}{*}{TQA} & \multicolumn{1}{l|}{Contriever} & 69.7 & 75.5 & \textbf{72.5}  \\ \hline
\end{tabular}
\end{adjustbox}
\caption{Exact Match scores using different retrievers on Natural Questions and TriviaQA development sets.
}
\label{tab:IncrementalInferenceResult}
\end{table}

Using the probing patterns established in Section~\ref{sec:psgtype}, we achieve significantly higher EM scores with fewer passages by only using the positive-leaning passages (Table~\ref{tab:IncrementalInferenceResult}). Removing \textbf{DN} and \textbf{SN} clearly demonstrates redudancy of these passages where EM@100 for probe 3 approaches AcEM@100. Scores for all other probing patterns are reported in Appendix~\ref{sec:appendix:probe_pattern}. Comparison between the other probe types indicates that the model is typically not confident with \textbf{IZ}, and it is helpful to retain this information rather than discard it.

\begin{figure}[!t]
  \centering
  \includegraphics[width=0.6\linewidth]{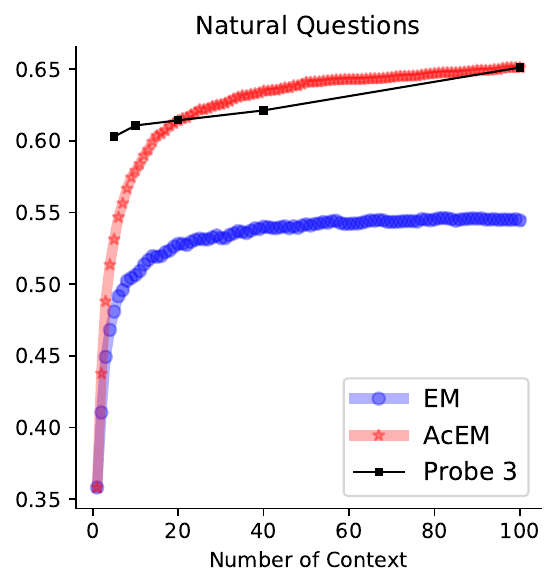}
  \caption{Probe 3 with varying size of input passages compared to incremental inference on DPR retrieved passages on Natural Questions test set. }
  \label{fig:nqtest}
\end{figure}

\begin{figure}[!t]
  \centering
  \includegraphics[width=0.6\linewidth]{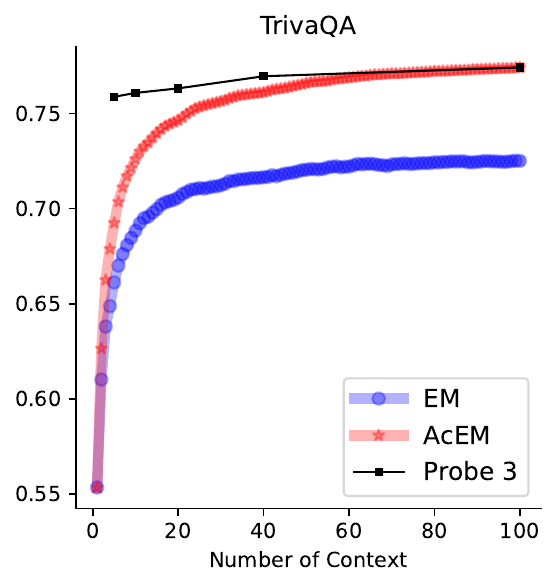}
  \caption{Probe 3 with varying size of input passages compared to incremental inference on DPR retrieved passages on TriviaQA test set.}
  \label{fig:triviaqatest}
\end{figure}

\paragraph{Few-sentence prediction}
We apply Probe 3 varying the number of passages from 5, 10, 20, 40 and 100 to evaluate whether we can attain higher accuracy while using a much smaller budget of passages for the reader model. With 5 passages, the model attains more than a 12\% increase in EM@5 for NQ and 8\% over EM@5 for TriviaQA on the test set. Results are reported in Figure~\ref{fig:nqtest} and Figure~\ref{fig:triviaqatest}.

While models can attain higher performance by adding more passages to their input, this comes at considerable computational costs, leading to increased memory usage and latency in prediction. Our results indicate that with just using 5 passages, state of the art EM scores can be exceeded with judicious filtering of the retrieved passages. It is crucial that retrieval models not only consider relevance for the end-user but also consider how irrelevant passages can cause a catastrophic interaction with the downstream reader model. 


\subsection{Attention Inference}
\label{result:attention-analysis}
\begin{figure}[!t]
  \centering
  \includegraphics[width=0.75\linewidth]{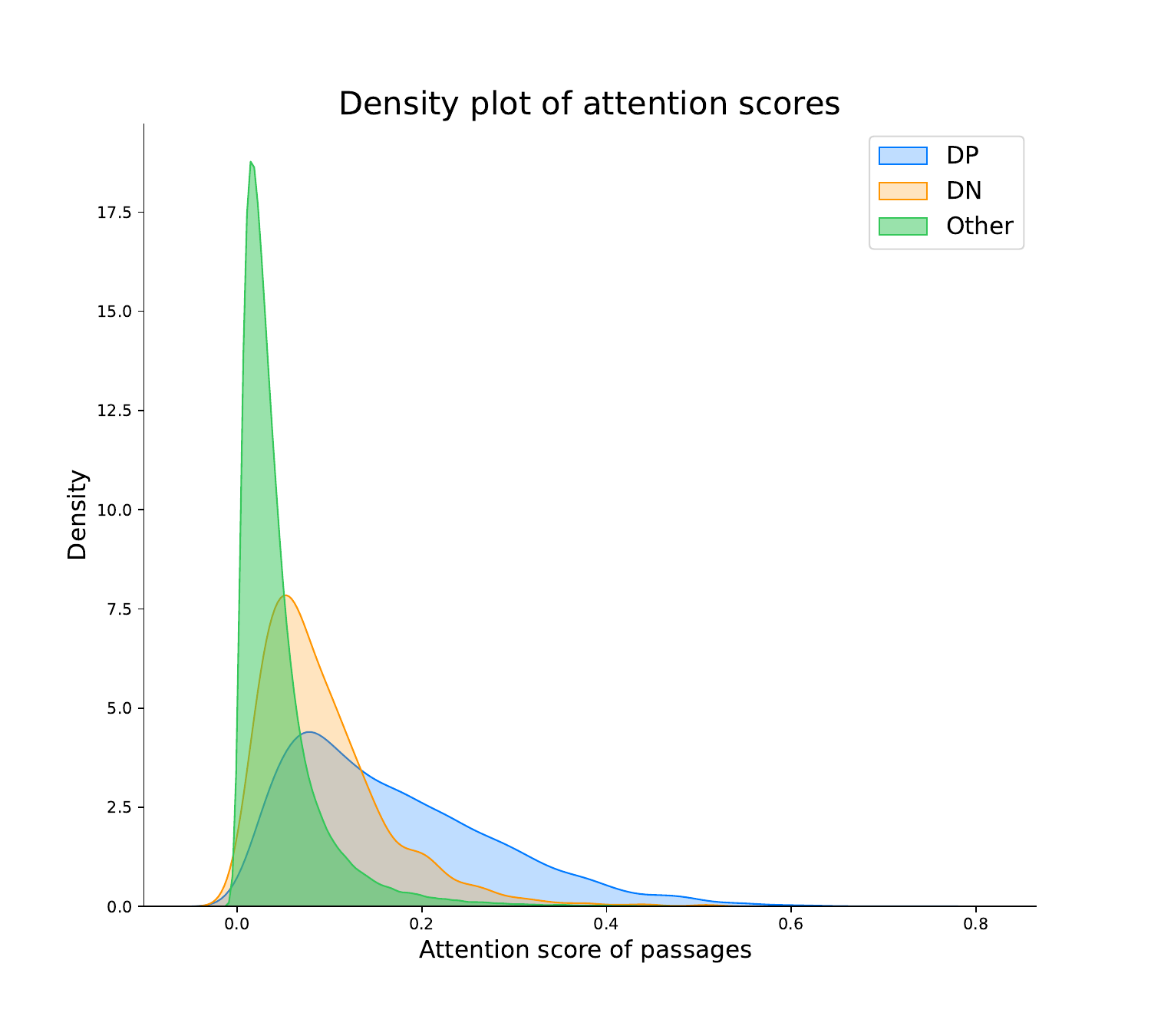}
  \caption{Attention density plot of top-20 passages in NQ development set. \textbf{DP} contexts present relatively high attention scores but cannot easily be disentangled.}
  \label{fig:attention-density}
\end{figure}

Figure~\ref{fig:attention-density} visualizes the distribution of \textbf{DP}, \textbf{DN}, and other passage types, showing a more dispersed and higher attention score distribution of \textbf{DP}, which aligns with the previous finding \cite{izacard2022distilling}. Interestingly, however, focusing on high attention scores results in a monotonic decline in performance, as shown in Table~\ref{tab:attention-analysis-result}. This implies that passages with high attention scores do not always guarantee the correct answer. In addition, we investigate the highest attention score in the presence of \textbf{DN}. Table~\ref{tab:attention-analysis-result2} illustrates \textbf{DP} accounts for 40 \% of the highest attention while \textbf{DN} only represents 26\%. However, 67 \% of transitioned answers are located within \textbf{DN}, while only 13 \% are attributed to \textbf{DP}, further highlighting an ambiguous relationship between attention and answers.

\begin{table}[t]
\small
\centering
\begin{adjustbox}{scale=0.9}
\begin{tabular}{|c|c|}
\hline
\textbf{Attention score threshold} & \textbf{Exact Match} \\
\hline
0.025 & 49.5 \\
0.05 & 49.11 \\
0.075 & 48.30 \\
0.1 & 47.55 \\
0.2 & 46.59 \\ \hline
AcEM@20 & 58.11 \\
EM@20  & 50.21 \\
\hline
\end{tabular}
\end{adjustbox}
\caption{EM score of various threshold values on top-20 passages. In cases where none of the contexts met the threshold values, all top-20 passages were used.}
\label{tab:attention-analysis-result}
\end{table}
\begin{table}[t]
\small
\centering
\begin{adjustbox}{scale=0.9}
\begin{tabular}{clcc|c}
\hline
\textbf{Feature}               & \multicolumn{1}{|c|}{\textbf{Context Types}}   & \textbf{Percentage}  \\ \hline
                      & \multicolumn{1}{|c|}{\textbf{DP}}        & 40.6    \\
\multirow{-2}{*}{Highest Attention Score}  & \multicolumn{1}{|c|}{\textbf{DN}}  & 26.3 \\ \hline
                      & \multicolumn{1}{|c|}{\textbf{DP}}        & 13.6    \\
\multirow{-2}{*}{Transformed Prediction} & \multicolumn{1}{|c|}{\textbf{DN}} & 67.7   \\ \hline
\end{tabular}
\end{adjustbox}
\caption{Out of \textbf{DN} 1091 cases, 40.6 \% \textbf{DP} still exhibits the highest attentions score, but 67.7 \% of transitioned predictions comes from \textbf{DN}.
}
\label{tab:attention-analysis-result2}
\end{table}

\subsection{Binary Inference}
\label{result:binary-analysis}
\begin{table}[t]
\small
\centering
\begin{adjustbox}{scale=0.8}
\begin{tabular}{c|c|c|c|c}
\hline
\textbf{Model} & \textbf{Acc} & \textbf{Pre} & \textbf{Re} & \textbf{F-1} \\ \hline
RoBERTAa (Large) & 0.76 & 0.76 & 1.0 & 0.76\\
FiD Encoder (Large) & 0.74 & 0.77 & 0.94 & 0.85 \\
T5 Encoder(Large) & 0.75 & 0.76 & 0.987 & 0.86 \\
\hline
\end{tabular}
\end{adjustbox}
\caption{Binary classification results of different models on the development set. All three models are highly likey to predict \textbf{DP}s.}
\label{tab:binary-models}
\end{table}
We observe that it is challenging for models to differentiate between damaging contexts and positive ones. 
Table~\ref{tab:binary-models} presents the results of binary classifications. No model trained well: all models exhibit a high recall rate at the expense of precision, suggesting a strong tendency to erroneously classify most passages as \textbf{DP}s. We conjecture that the retrieved passagess already demonstrate a high relevance score over the query, so model cannot differentiate them properly.

\section{Ablations}

\subsection{Passage Type Classification}
\label{ablation:end2e}
Binary classification of passage type as \textbf{DP} or \textbf{DN} (Section~\ref{result:binary-analysis}) was not infomrative. We further consider passage classification as a multi-class classification on all context types and evaluate its end-to-end performance. We use the same models and datasets as in Section~\ref{exp:binary}, resulting in 700,500 and 175,200 instances for training and evaluation, respectively. For the end-to-end evaluation, we employ the NQ test set. All training is performed with a batch size of 64, a learning rate of 5e-5, 11,000 steps (approximately 1 epoch). Multi-class classification results indicate that all models struggle with distinguishing between \textbf{DP} and \textbf{DN}, favoring predictions of \textbf{SP}, which constitutes 50\% of the instances. Detailed results are reported in Appendix~\ref{appen:multi-class} which show that all probe methods exhibited either lower or equal performance compared to EM@100, falling significantly below Probe 3. Both ablations underscore the challenge in models for predicting EM patterns and passage types, reflecting the difficulty in distinguishing \textbf{DP} and \textbf{DN} within the retrieved contexts.

\subsection{Qualitative Analysis on Definite Negative}
\label{ablation:qualitative-analysis}
Results from Section~\ref{sec:selection-inference} indicate that removing \textbf{DN} and \textbf{SN} passage types can significantly enhance the EM score. However, this approach relies on the availability of known answers to identify the damaging passage types. 
To discern patterns in the transition of EM patterns, we manually examine instances containing \textbf{DN} passages. We sample 100 examples for NQ development set and discover that out of these instances, 51 cases identified as correct answers, misled by dataset issues such as \textbf{Equivalent answers}, where answers that are semantically equivalent but have variations on surface form (e.g. alternative spellings), \textbf{Alternative answer} where only one answer out of a set is labelled in the dataset (e.g. multiple actors play a character), and \textbf{Temporality} where the correct answer depends on the current context (e.g. questions asking about latest events).

This finding supports \citealt{bulian-etal-2022-tomayto} that the EM may not fully capture the impact of issues like \textbf{Equivalent answer}. The remaining 49 cases are indeed affected by \textbf{DN}. This analysis highlights the limitations of inferring damaging contexts without answers. For more details and examples of qualitative analysis on \textbf{DN}, please refer to Appendix~\ref{append:qualitative-analysis}.

\subsection{Semantically Equivalent Answers}
\label{ablation:8-1-2_semantic}
Contemporaneous work from \citealt{kamalloo-etal-2023-evaluating} asserts the limitation of a lexical matching system (EM score) that leads to the underestimation of Reader's performance. They demonstrate that prompting LLM to assess the final outcome yields similar results to manual evaluation. Inspired by this, we re-evaluate our results via zero-shot prompting, following the methodology outlined in \citealt{kamalloo-etal-2023-evaluating}. We calculate adjusted EM and AcEM, which refer to as SeEM(Semantic Equivalent EM) and SeAcEM(Semantic Equivalent AcEM), to assess the damaging effects. Our experiments are conducted on a subset of NQ dev set (1752 instances) using \textbf{Claude2} and \textbf{gpt-3.5-turbo} (detailed in Appendix~\ref{appen:sem_em}).

\begin{figure}[t]
\centering
    \begin{subfigure}[h]{0.49\linewidth}
        \includegraphics[width=\linewidth]{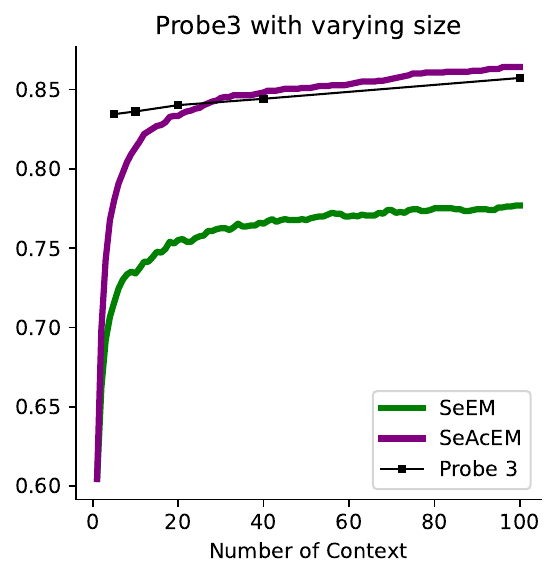}
        \caption{\textbf{gpt-3.5-turbo}}
        \label{fig:sem-probe3-fig1}
    \end{subfigure}
        \begin{subfigure}[h]{0.49\linewidth}
        \includegraphics[width=\linewidth]{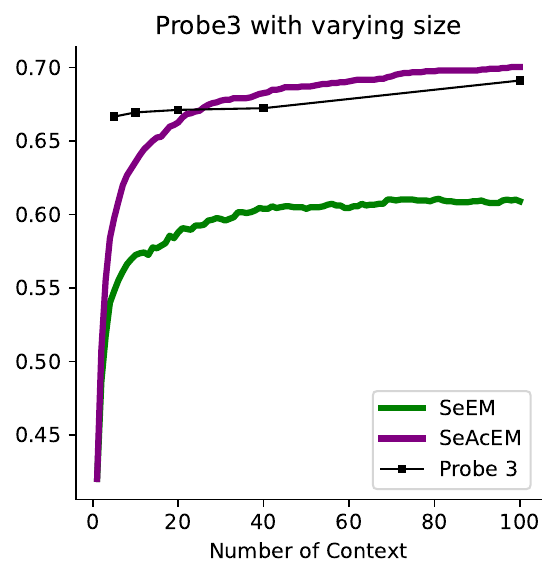}
        \caption{\textbf{Claude2}}
        \label{fig:sem-probe3-fig2}
    \end{subfigure}
\caption{Probe3 evaluations with varying sizes(5, 10, 20, 40, 100) on NQ dev subsets. Probe3 can achieve a performance close to the SeAcEM@100 with 20 times fewer contexts.}
\label{fig:sem-probe-equi}
\end{figure}

Figure~\ref{fig:sem-equi} reports the results of SeEM and SeAcEM in comparison to EM and AcEM. Increase in performance were observed for both \textbf{Claude2} and \textbf{gpt-3.5-turbo} to evaluate semantic equivalence. 
There was a 25.6\% increase in the \textbf{gpt-3.5-turbo} setting (EM@100: 0.520 to SeEM@100: 0.776) and an 8.9\% increase in the \textbf{Claude2} setting (EM@100: 0.520 to SeEM@100: 0.609). However, discrepancies of 8.8\% and 9.1\% persist between SeAcEM@100 and SeEM@100 in both settings, indicating that damaging effects still remain despite the semantic equivalent adjustments. To evaluate the effectiveness of our Probe3, we apply the same procedure outlined in Section~\ref{sec:selection-inference}. The results reported in Figure~\ref{fig:sem-probe-equi} demonstrate a 5.8\% and 5.7\% increase Probe3@5 (0.834, 0.666) compared to the conventional approach SeEM@100 (0.776, 0.609). Notably, this improvement is achieved using 1/20th of the context. This result illustrates the persistent presence of damaging passages even after adjusting for semantic equivalence, emphasizing the need for filtering out damaging passages.

\begin{figure}[t]
\centering
    \begin{subfigure}[h]{0.49\linewidth}
        \includegraphics[width=\linewidth]{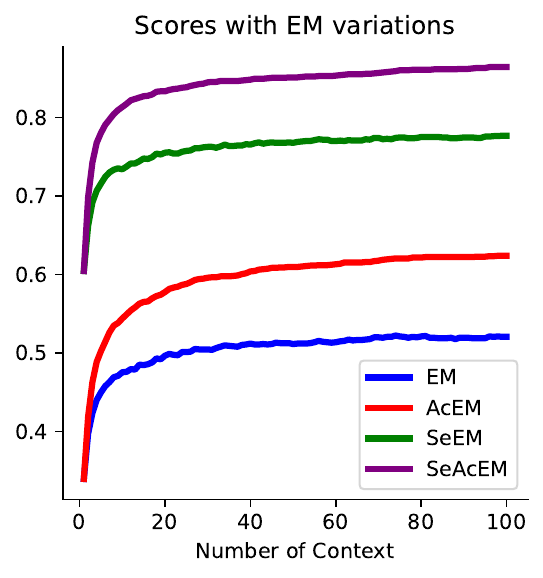}
        \caption{\textbf{gpt-3.5-turbo}}
        \label{fig:sem-fig1}
    \end{subfigure}
        \begin{subfigure}[h]{0.49\linewidth}
        \includegraphics[width=\linewidth]{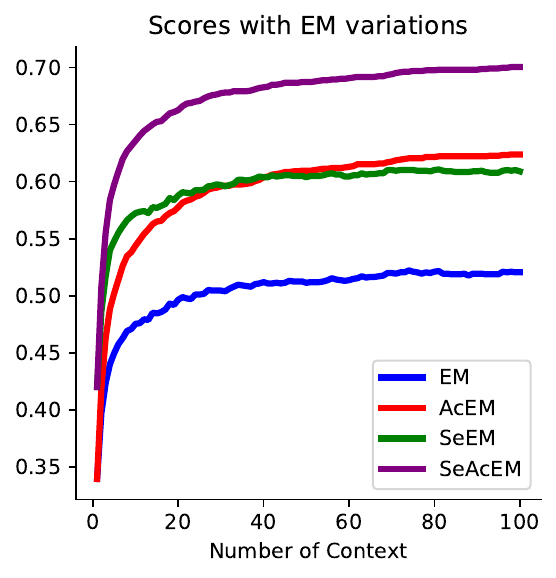}
        \caption{\textbf{Claude2}}
        \label{fig:sem-fig2}
    \end{subfigure}
\caption{Comparison of different metrics on NQ dev subsets, comprising 1752 instances. SeEM and AcEM are adjusted versions of EM, correcting semantically equivalent answers via LLM prompting.}
\label{fig:sem-equi}
\end{figure}

\section{Conclusions}
The reader models in retrieve-then-read pipelines are sensitive to the retrieved contexts when generating answers. Damaging passages in this set can lead to incorrect responses. Filtering damaging passages results in increases in EM scores without the need for architectural modifications. Despite shortcomings in evaluating QA with exact match, we demonstrate that by filtering passages, models can achieve 10\% higher EM scores using subsets of context that are 20X times smaller. 

\section{Limitations}
Identifying the behavior of black-box models is challenging. While we identify different subsets of evidence that the model reacts well to when generating a correct answer, there is no guarantee that these subsets of information correspond to what humans users would consider useful. Furthermore, some of the reasons the model changed its prediction, such as generating a more specific answer, would be correct if multiple references were available for evaluating the models. However, these alternative answers are not available in the datasets, which means we are optimizing the models for a limited subset of truly valid answers. Lastly, our approach may not be practical for decoder-only LLMs where the order of context/answer choices varies the outcome. To assess the answerability of the given $n$ candidate contexts, $\mathcal{O}(n!)$ inferences are required for LLMs, while FiD only needs one inference due to its order-invariance property.

We report a limitation, evaluation and position of established modelling techniques that can help guide the community for future research. If models can effectively leverage external information, they should be capable of using text as an interpretable source of information rather than relying solely on knowledge that is stored within inaccessible model parameters. This approach may contribute to a future with NLP models that are more interpretable and controllable.

\section{Acknowledgements}
This work was supported by Institute of Information \& communications Technology Planning \& Evaluation (IITP) grant funded by the Korea government (MSIT) (No.2019-0-00075, Artificial Intelligence Graduate School Program (KAIST)) and Artificial intelligence industrial convergence cluster development project funded by the Ministry of Science and ICT(MSIT, Korea) \& Gwangju Metropolitan City.

\bibliography{anthology,custom}

\begin{thebibliography}{40}
\expandafter\ifx\csname natexlab\endcsname\relax\def\natexlab#1{#1}\fi

\bibitem[{Asai et~al.(2021)Asai, Gardner, and Hajishirzi}]{DBLP:journals/corr/abs-2112-08688}
Akari Asai, Matt Gardner, and Hannaneh Hajishirzi. 2021.
\newblock \href {http://arxiv.org/abs/2112.08688} {Evidentiality-guided generation for knowledge-intensive {NLP} tasks}.
\newblock \emph{CoRR}, abs/2112.08688.

\bibitem[{Bevilacqua et~al.(2022)Bevilacqua, Ottaviano, Lewis, tau Yih, Riedel, and Petroni}]{bevilacqua2022autoregressive}
Michele Bevilacqua, Giuseppe Ottaviano, Patrick Lewis, Wen tau Yih, Sebastian Riedel, and Fabio Petroni. 2022.
\newblock \href {https://arxiv.org/abs/2204.10628} {Autoregressive search engines: Generating substrings as document identifiers}.
\newblock In \emph{arXiv pre-print 2204.10628}.

\bibitem[{Bulian et~al.(2022)Bulian, Buck, Gajewski, B{\"o}rschinger, and Schuster}]{bulian-etal-2022-tomayto}
Jannis Bulian, Christian Buck, Wojciech Gajewski, Benjamin B{\"o}rschinger, and Tal Schuster. 2022.
\newblock \href {https://aclanthology.org/2022.emnlp-main.20} {Tomayto, tomahto. beyond token-level answer equivalence for question answering evaluation}.
\newblock In \emph{Proceedings of the 2022 Conference on Empirical Methods in Natural Language Processing}, pages 291--305, Abu Dhabi, United Arab Emirates. Association for Computational Linguistics.

\bibitem[{Cao et~al.(2020)Cao, Izacard, Riedel, and Petroni}]{DBLP:journals/corr/abs-2010-00904}
Nicola~De Cao, Gautier Izacard, Sebastian Riedel, and Fabio Petroni. 2020.
\newblock \href {http://arxiv.org/abs/2010.00904} {Autoregressive entity retrieval}.
\newblock \emph{CoRR}, abs/2010.00904.

\bibitem[{Chen et~al.(2017)Chen, Fisch, Weston, and Bordes}]{chen-etal-2017-reading}
Danqi Chen, Adam Fisch, Jason Weston, and Antoine Bordes. 2017.
\newblock \href {https://doi.org/10.18653/v1/P17-1171} {Reading {W}ikipedia to answer open-domain questions}.
\newblock In \emph{Proceedings of the 55th Annual Meeting of the Association for Computational Linguistics (Volume 1: Long Papers)}, pages 1870--1879, Vancouver, Canada. Association for Computational Linguistics.

\bibitem[{de~Jong et~al.(2022)de~Jong, Zemlyanskiy, Ainslie, FitzGerald, Sanghai, Sha, and Cohen}]{https://doi.org/10.48550/arxiv.2212.08153}
Michiel de~Jong, Yury Zemlyanskiy, Joshua Ainslie, Nicholas FitzGerald, Sumit Sanghai, Fei Sha, and William Cohen. 2022.
\newblock \href {https://doi.org/10.48550/ARXIV.2212.08153} {Fido: Fusion-in-decoder optimized for stronger performance and faster inference}.

\bibitem[{Devlin et~al.(2018)Devlin, Chang, Lee, and Toutanova}]{DBLP:journals/corr/abs-1810-04805}
Jacob Devlin, Ming{-}Wei Chang, Kenton Lee, and Kristina Toutanova. 2018.
\newblock \href {http://arxiv.org/abs/1810.04805} {{BERT:} pre-training of deep bidirectional transformers for language understanding}.
\newblock \emph{CoRR}, abs/1810.04805.

\bibitem[{Ferrante et~al.(2018)Ferrante, Ferro, and Pontarollo}]{ferrante2018general}
Marco Ferrante, Nicola Ferro, and Silvia Pontarollo. 2018.
\newblock A general theory of ir evaluation measures.
\newblock \emph{IEEE Transactions on Knowledge and Data Engineering}, 31(3):409--422.

\bibitem[{Glass et~al.(2022)Glass, Rossiello, Chowdhury, Naik, Cai, and Gliozzo}]{glass-etal-2022-re2g}
Michael Glass, Gaetano Rossiello, Md~Faisal~Mahbub Chowdhury, Ankita Naik, Pengshan Cai, and Alfio Gliozzo. 2022.
\newblock \href {https://doi.org/10.18653/v1/2022.naacl-main.194} {{R}e2{G}: Retrieve, rerank, generate}.
\newblock In \emph{Proceedings of the 2022 Conference of the North American Chapter of the Association for Computational Linguistics: Human Language Technologies}, pages 2701--2715, Seattle, United States. Association for Computational Linguistics.

\bibitem[{Hofst{\"a}tter et~al.(2022)Hofst{\"a}tter, Chen, Raman, and Zamani}]{hofstatter2022fid}
Sebastian Hofst{\"a}tter, Jiecao Chen, Karthik Raman, and Hamed Zamani. 2022.
\newblock Fid-light: Efficient and effective retrieval-augmented text generation.
\newblock \emph{arXiv preprint arXiv:2209.14290}.

\bibitem[{Iyer et~al.(2021)Iyer, Min, Mehdad, and Yih}]{iyer2020reconsider}
Srinivasan Iyer, Sewon Min, Yashar Mehdad, and Wen-tau Yih. 2021.
\newblock Reconsider: Re-ranking using span-focused cross-attention for open domain question answering.
\newblock In \emph{NAACL}.

\bibitem[{Izacard et~al.(2021)Izacard, Caron, Hosseini, Riedel, Bojanowski, Joulin, and Grave}]{izacard2021contriever}
Gautier Izacard, Mathilde Caron, Lucas Hosseini, Sebastian Riedel, Piotr Bojanowski, Armand Joulin, and Edouard Grave. 2021.
\newblock \href {https://doi.org/10.48550/ARXIV.2112.09118} {Unsupervised dense information retrieval with contrastive learning}.

\bibitem[{Izacard and Grave(2020)}]{izacard2020leveraging}
Gautier Izacard and Edouard Grave. 2020.
\newblock \href {https://arxiv.org/abs/2007.0128} {Leveraging passage retrieval with generative models for open domain question answering}.

\bibitem[{Izacard and Grave(2022)}]{izacard2022distilling}
Gautier Izacard and Edouard Grave. 2022.
\newblock \href {http://arxiv.org/abs/2012.04584} {Distilling knowledge from reader to retriever for question answering}.

\bibitem[{Izacard et~al.(2022)Izacard, Lewis, Lomeli, Hosseini, Petroni, Schick, Dwivedi-Yu, Joulin, Riedel, and Grave}]{izacard2022few}
Gautier Izacard, Patrick Lewis, Maria Lomeli, Lucas Hosseini, Fabio Petroni, Timo Schick, Jane Dwivedi-Yu, Armand Joulin, Sebastian Riedel, and Edouard Grave. 2022.
\newblock Few-shot learning with retrieval augmented language models.
\newblock \emph{arXiv preprint arXiv:2208.03299}.

\bibitem[{Jiang et~al.(2020)Jiang, Bordia, Zhong, Dognin, Singh, and Bansal}]{jiang-etal-2020-hover}
Yichen Jiang, Shikha Bordia, Zheng Zhong, Charles Dognin, Maneesh Singh, and Mohit Bansal. 2020.
\newblock \href {https://doi.org/10.18653/v1/2020.findings-emnlp.309} {{H}o{V}er: A dataset for many-hop fact extraction and claim verification}.
\newblock In \emph{Findings of the Association for Computational Linguistics: EMNLP 2020}, pages 3441--3460, Online. Association for Computational Linguistics.

\bibitem[{Joshi et~al.(2017)Joshi, Choi, Weld, and Zettlemoyer}]{joshi-etal-2017-triviaqa}
Mandar Joshi, Eunsol Choi, Daniel Weld, and Luke Zettlemoyer. 2017.
\newblock \href {https://doi.org/10.18653/v1/P17-1147} {{T}rivia{QA}: A large scale distantly supervised challenge dataset for reading comprehension}.
\newblock In \emph{Proceedings of the 55th Annual Meeting of the Association for Computational Linguistics (Volume 1: Long Papers)}, pages 1601--1611, Vancouver, Canada. Association for Computational Linguistics.

\bibitem[{Kamalloo et~al.(2023)Kamalloo, Dziri, Clarke, and Rafiei}]{kamalloo-etal-2023-evaluating}
Ehsan Kamalloo, Nouha Dziri, Charles Clarke, and Davood Rafiei. 2023.
\newblock \href {https://doi.org/10.18653/v1/2023.acl-long.307} {Evaluating open-domain question answering in the era of large language models}.
\newblock In \emph{Proceedings of the 61st Annual Meeting of the Association for Computational Linguistics (Volume 1: Long Papers)}, pages 5591--5606, Toronto, Canada. Association for Computational Linguistics.

\bibitem[{Karpukhin et~al.(2020)Karpukhin, Oguz, Min, Lewis, Wu, Edunov, Chen, and Yih}]{karpukhin-etal-2020-dense}
Vladimir Karpukhin, Barlas Oguz, Sewon Min, Patrick Lewis, Ledell Wu, Sergey Edunov, Danqi Chen, and Wen-tau Yih. 2020.
\newblock \href {https://doi.org/10.18653/v1/2020.emnlp-main.550} {Dense passage retrieval for open-domain question answering}.
\newblock In \emph{Proceedings of the 2020 Conference on Empirical Methods in Natural Language Processing (EMNLP)}, pages 6769--6781, Online. Association for Computational Linguistics.

\bibitem[{Kedia et~al.(2022)Kedia, Zaidi, and Lee}]{https://doi.org/10.48550/arxiv.2211.10147}
Akhil Kedia, Mohd~Abbas Zaidi, and Haejun Lee. 2022.
\newblock \href {https://doi.org/10.48550/ARXIV.2211.10147} {Fie: Building a global probability space by leveraging early fusion in encoder for open-domain question answering}.

\bibitem[{Kongyoung et~al.(2022)Kongyoung, Macdonald, and Ounis}]{kongyoung-etal-2022-monoqa}
Sarawoot Kongyoung, Craig Macdonald, and Iadh Ounis. 2022.
\newblock \href {https://doi.org/10.18653/v1/2022.emnlp-main.485} {mono{QA}: Multi-task learning of reranking and answer extraction for open-retrieval conversational question answering}.
\newblock In \emph{Proceedings of the 2022 Conference on Empirical Methods in Natural Language Processing}, pages 7207--7218, Abu Dhabi, United Arab Emirates. Association for Computational Linguistics.

\bibitem[{Krishna et~al.(2021)Krishna, Riedel, and Vlachos}]{DBLP:journals/corr/abs-2108-11357}
Amrith Krishna, Sebastian Riedel, and Andreas Vlachos. 2021.
\newblock \href {http://arxiv.org/abs/2108.11357} {Proofver: Natural logic theorem proving for fact verification}.
\newblock \emph{CoRR}, abs/2108.11357.

\bibitem[{Kwiatkowski et~al.(2019)Kwiatkowski, Palomaki, Redfield, Collins, Parikh, Alberti, Epstein, Polosukhin, Devlin, Lee, Toutanova, Jones, Kelcey, Chang, Dai, Uszkoreit, Le, and Petrov}]{kwiatkowski-etal-2019-natural}
Tom Kwiatkowski, Jennimaria Palomaki, Olivia Redfield, Michael Collins, Ankur Parikh, Chris Alberti, Danielle Epstein, Illia Polosukhin, Jacob Devlin, Kenton Lee, Kristina Toutanova, Llion Jones, Matthew Kelcey, Ming-Wei Chang, Andrew~M. Dai, Jakob Uszkoreit, Quoc Le, and Slav Petrov. 2019.
\newblock \href {https://doi.org/10.1162/tacl_a_00276} {Natural questions: A benchmark for question answering research}.
\newblock \emph{Transactions of the Association for Computational Linguistics}, 7:452--466.

\bibitem[{Lee et~al.(2019)Lee, Chang, and Toutanova}]{DBLP:journals/corr/abs-1906-00300}
Kenton Lee, Ming{-}Wei Chang, and Kristina Toutanova. 2019.
\newblock \href {http://arxiv.org/abs/1906.00300} {Latent retrieval for weakly supervised open domain question answering}.
\newblock \emph{CoRR}, abs/1906.00300.

\bibitem[{Levy et~al.(2017)Levy, Seo, Choi, and Zettlemoyer}]{levy-etal-2017-zero}
Omer Levy, Minjoon Seo, Eunsol Choi, and Luke Zettlemoyer. 2017.
\newblock \href {https://doi.org/10.18653/v1/K17-1034} {Zero-shot relation extraction via reading comprehension}.
\newblock In \emph{Proceedings of the 21st Conference on Computational Natural Language Learning ({C}o{NLL} 2017)}, pages 333--342, Vancouver, Canada. Association for Computational Linguistics.

\bibitem[{Lewis et~al.(2020)Lewis, Perez, Piktus, Petroni, Karpukhin, Goyal, K\"{u}ttler, Lewis, Yih, Rockt\"{a}schel, Riedel, and Kiela}]{NEURIPS2020_6b493230}
Patrick Lewis, Ethan Perez, Aleksandra Piktus, Fabio Petroni, Vladimir Karpukhin, Naman Goyal, Heinrich K\"{u}ttler, Mike Lewis, Wen-tau Yih, Tim Rockt\"{a}schel, Sebastian Riedel, and Douwe Kiela. 2020.
\newblock \href {https://proceedings.neurips.cc/paper_files/paper/2020/file/6b493230205f780e1bc26945df7481e5-Paper.pdf} {Retrieval-augmented generation for knowledge-intensive nlp tasks}.
\newblock In \emph{Advances in Neural Information Processing Systems}, volume~33, pages 9459--9474. Curran Associates, Inc.

\bibitem[{Liang et~al.(2022)Liang, Bommasani, Lee, Tsipras, Soylu, Yasunaga, Zhang, Narayanan, Wu, Kumar, Newman, Yuan, Yan, Zhang, Cosgrove, Manning, Ré, Acosta-Navas, Hudson, Zelikman, Durmus, Ladhak, Rong, Ren, Yao, Wang, Santhanam, Orr, Zheng, Yuksekgonul, Suzgun, Kim, Guha, Chatterji, Khattab, Henderson, Huang, Chi, Xie, Santurkar, Ganguli, Hashimoto, Icard, Zhang, Chaudhary, Wang, Li, Mai, Zhang, and Koreeda}]{liang2022holistic}
Percy Liang, Rishi Bommasani, Tony Lee, Dimitris Tsipras, Dilara Soylu, Michihiro Yasunaga, Yian Zhang, Deepak Narayanan, Yuhuai Wu, Ananya Kumar, Benjamin Newman, Binhang Yuan, Bobby Yan, Ce~Zhang, Christian Cosgrove, Christopher~D. Manning, Christopher Ré, Diana Acosta-Navas, Drew~A. Hudson, Eric Zelikman, Esin Durmus, Faisal Ladhak, Frieda Rong, Hongyu Ren, Huaxiu Yao, Jue Wang, Keshav Santhanam, Laurel Orr, Lucia Zheng, Mert Yuksekgonul, Mirac Suzgun, Nathan Kim, Neel Guha, Niladri Chatterji, Omar Khattab, Peter Henderson, Qian Huang, Ryan Chi, Sang~Michael Xie, Shibani Santurkar, Surya Ganguli, Tatsunori Hashimoto, Thomas Icard, Tianyi Zhang, Vishrav Chaudhary, William Wang, Xuechen Li, Yifan Mai, Yuhui Zhang, and Yuta Koreeda. 2022.
\newblock \href {http://arxiv.org/abs/2211.09110} {Holistic evaluation of language models}.

\bibitem[{Liu et~al.(2023)Liu, Lin, Hewitt, Paranjape, Bevilacqua, Petroni, and Liang}]{liu2023lost}
Nelson~F. Liu, Kevin Lin, John Hewitt, Ashwin Paranjape, Michele Bevilacqua, Fabio Petroni, and Percy Liang. 2023.
\newblock \href {http://arxiv.org/abs/2307.03172} {Lost in the middle: How language models use long contexts}.

\bibitem[{Lyu et~al.(2023)Lyu, Grafberger, Biegel, Wei, Cao, Schelter, and Zhang}]{lyu2023improving}
Xiaozhong Lyu, Stefan Grafberger, Samantha Biegel, Shaopeng Wei, Meng Cao, Sebastian Schelter, and Ce~Zhang. 2023.
\newblock \href {http://arxiv.org/abs/2307.03027} {Improving retrieval-augmented large language models via data importance learning}.

\bibitem[{Petroni et~al.(2021)Petroni, Piktus, Fan, Lewis, Yazdani, De~Cao, Thorne, Jernite, Karpukhin, Maillard, Plachouras, Rockt{\"a}schel, and Riedel}]{petroni-etal-2021-kilt}
Fabio Petroni, Aleksandra Piktus, Angela Fan, Patrick Lewis, Majid Yazdani, Nicola De~Cao, James Thorne, Yacine Jernite, Vladimir Karpukhin, Jean Maillard, Vassilis Plachouras, Tim Rockt{\"a}schel, and Sebastian Riedel. 2021.
\newblock \href {https://doi.org/10.18653/v1/2021.naacl-main.200} {{KILT}: a benchmark for knowledge intensive language tasks}.
\newblock In \emph{Proceedings of the 2021 Conference of the North American Chapter of the Association for Computational Linguistics: Human Language Technologies}, pages 2523--2544, Online. Association for Computational Linguistics.

\bibitem[{Raffel et~al.(2019)Raffel, Shazeer, Roberts, Lee, Narang, Matena, Zhou, Li, and Liu}]{DBLP:journals/corr/abs-1910-10683}
Colin Raffel, Noam Shazeer, Adam Roberts, Katherine Lee, Sharan Narang, Michael Matena, Yanqi Zhou, Wei Li, and Peter~J. Liu. 2019.
\newblock \href {http://arxiv.org/abs/1910.10683} {Exploring the limits of transfer learning with a unified text-to-text transformer}.
\newblock \emph{CoRR}, abs/1910.10683.

\bibitem[{Rajpurkar et~al.(2016)Rajpurkar, Zhang, Lopyrev, and Liang}]{rajpurkar-etal-2016-squad}
Pranav Rajpurkar, Jian Zhang, Konstantin Lopyrev, and Percy Liang. 2016.
\newblock \href {https://doi.org/10.18653/v1/D16-1264} {{SQ}u{AD}: 100,000+ questions for machine comprehension of text}.
\newblock In \emph{Proceedings of the 2016 Conference on Empirical Methods in Natural Language Processing}, pages 2383--2392, Austin, Texas. Association for Computational Linguistics.

\bibitem[{Sauchuk et~al.(2022)Sauchuk, Thorne, Halevy, Tonellotto, and Silvestri}]{10.1145/3477495.3532034}
Artsiom Sauchuk, James Thorne, Alon Halevy, Nicola Tonellotto, and Fabrizio Silvestri. 2022.
\newblock \href {https://doi.org/10.1145/3477495.3532034} {On the role of relevance in natural language processing tasks}.
\newblock In \emph{Proceedings of the 45th International ACM SIGIR Conference on Research and Development in Information Retrieval}, SIGIR '22, page 1785–1789, New York, NY, USA. Association for Computing Machinery.

\bibitem[{Terra and Warren(2005)}]{10.1145/1099554.1099646}
Egidio Terra and Robert Warren. 2005.
\newblock \href {https://doi.org/10.1145/1099554.1099646} {Poison pills: Harmful relevant documents in feedback}.
\newblock In \emph{Proceedings of the 14th ACM International Conference on Information and Knowledge Management}, CIKM '05, page 319–320, New York, NY, USA. Association for Computing Machinery.

\bibitem[{Thakur et~al.(2021)Thakur, Reimers, R{\"u}ckl{\'e}, Srivastava, and Gurevych}]{thakur2021beir}
Nandan Thakur, Nils Reimers, Andreas R{\"u}ckl{\'e}, Abhishek Srivastava, and Iryna Gurevych. 2021.
\newblock Beir: A heterogenous benchmark for zero-shot evaluation of information retrieval models.
\newblock \emph{arXiv preprint arXiv:2104.08663}.

\bibitem[{Thorne et~al.(2018)Thorne, Vlachos, Christodoulopoulos, and Mittal}]{thorne-etal-2018-fever}
James Thorne, Andreas Vlachos, Christos Christodoulopoulos, and Arpit Mittal. 2018.
\newblock \href {https://doi.org/10.18653/v1/N18-1074} {{FEVER}: a large-scale dataset for fact extraction and {VER}ification}.
\newblock In \emph{Proceedings of the 2018 Conference of the North {A}merican Chapter of the Association for Computational Linguistics: Human Language Technologies, Volume 1 (Long Papers)}, pages 809--819, New Orleans, Louisiana. Association for Computational Linguistics.

\bibitem[{Xu et~al.(2021)Xu, Liang, Huang, and Xiang}]{xu2021attentionguided}
Peng Xu, Davis Liang, Zhiheng Huang, and Bing Xiang. 2021.
\newblock \href {http://arxiv.org/abs/2110.06393} {Attention-guided generative models for extractive question answering}.

\bibitem[{Yang et~al.(2018)Yang, Qi, Zhang, Bengio, Cohen, Salakhutdinov, and Manning}]{yang-etal-2018-hotpotqa}
Zhilin Yang, Peng Qi, Saizheng Zhang, Yoshua Bengio, William Cohen, Ruslan Salakhutdinov, and Christopher~D. Manning. 2018.
\newblock \href {https://doi.org/10.18653/v1/D18-1259} {{H}otpot{QA}: A dataset for diverse, explainable multi-hop question answering}.
\newblock In \emph{Proceedings of the 2018 Conference on Empirical Methods in Natural Language Processing}, pages 2369--2380, Brussels, Belgium. Association for Computational Linguistics.

\bibitem[{Yu et~al.(2023)Yu, Iter, Wang, Xu, Ju, Sanyal, Zhu, Zeng, and Jiang}]{yu2023generate}
Wenhao Yu, Dan Iter, Shuohang Wang, Yichong Xu, Mingxuan Ju, Soumya Sanyal, Chenguang Zhu, Michael Zeng, and Meng Jiang. 2023.
\newblock Generate rather than retrieve: Large language models are strong context generators.
\newblock In \emph{International Conference for Learning Representation (ICLR)}.

\bibitem[{Zheng et~al.(2023)Zheng, Zhou, Meng, Zhou, and Huang}]{zheng2023large}
Chujie Zheng, Hao Zhou, Fandong Meng, Jie Zhou, and Minlie Huang. 2023.
\newblock \href {http://arxiv.org/abs/2309.03882} {On large language models' selection bias in multi-choice questions}.

\end{thebibliography}
\bibliographystyle{acl_natbib}

\clearpage
\onecolumn
\appendix

\section{Appendix}
\label{sec:appendix}
\subsection{Probe pattern performance}\label{sec:appendix:probe_pattern}
We report the performance of all probe patterns from Section~\ref{sec:probing}. Probe pattern 3 gives EM scores that are closest to the established AcEM@100 upper bound. 
\begin{table}[h!]
\small
\centering
\begin{tabular}{clcc|cccccc}
\hline
Dataset               & \multicolumn{1}{l|}{Retriever}   & EM@100 & AcEM@100 & Probe 1 & Probe 2 & Probe 3                               & Probe 4 & Probe 5 & Probe 6 \\ \hline
                      & \multicolumn{1}{l|}{DPR}        & 52.5 & 62.3 & 58.1   & 56.3   & \underline{\textbf{61.8}} & 61.7   & 55.2   & 56.0   \\
                      & \multicolumn{1}{l|}{SEAL}       & 50.0 & 59.4 & 43.9   & 30.0   & \textbf{52.9}                        & 44.9   & 52.5   & 44.6   \\
\multirow{-3}{*}{NQ}  & \multicolumn{1}{l|}{Contriever}  & 50.7 & 60.8 & 44.3   & 29.0   & \textbf{53.0}                        & 45.1   & 52.6   & 44.4   \\ \hline
                      & \multicolumn{1}{l|}{DPR}        & 72.3 & 77.7 & 76.2   & 75.1   & \underline{\textbf{77.6}} & 77.4    & 73.8   & 74.0   \\
                      & \multicolumn{1}{l|}{SEAL}       & 67.1 & 72.3 & 67.9   & 51.5   & \textbf{72.5}                        & 63.7   & 72.3   & 63.9   \\
\multirow{-3}{*}{TQA} & \multicolumn{1}{l|}{Contriever} & 69.7 & 75.5 & 68.4   & 51.7   & \textbf{72.5}                        & 62.9   & 72.3   & 62.9   \\ \hline
\end{tabular}
\caption{Exact Match scores using different retrievers on Natural Questions and TriviaQA development sets. The datasets retrieved by DPR exhibit a consistently high EM score across various Probes. This observation can be explained by the fact that the FiD models used for evaluation are trained on DPR retrieved datasets.
}
\label{tab:IncrementalInferenceResultAppendix}
\end{table}

\subsection{Difference between leave-one-out masking and EM pattern}
\label{appen:difference}
There are two key distinctions between \citealt{DBLP:journals/corr/abs-2112-08688} and our apporach. Firstly, while leave-one-out masking focuses on finding positive contexts, ours method(Passage Type Selection) aims to detect detrimental passages that can negatively impact the inference and it can also distinguish positive contexts effectively. Secondly, the time taken for inference to determine whether a passage is positive or negative is faster in our approach compared to leave-one-out masking. For instance, let $N-1$ be the number of contexts, $\mathcal{G}$ denote the reader model, and $\mathbf{P} = \{p_{1}, p_{2}, ..., p_{N-1} \}$ denote the set of retrieved passages for a given query. When new context, $p_{N}$, is added to the passages lists, it requires N inferences over N-1 passages in the case of leave-one-out masking, whereas ours requires a single inference over N passages. 

\subsection{Qualitative Analysis}
\label{append:qualitative-analysis}
In this section, we delve further into the qualitative analysis of \textbf{DN}. We focus on the top-20 passages as examining the top-100 passages would require a more extensive investigation. Out of the 8,757 cases in the NQ development set, 1,091 instances exhibit the presence of \textbf{DN}, accounting for approximately 12\% of the dataset. We selected a sample of the first 100 cases and discovered that out of these instances, 51 \textbf{DN} cases were attributed to limitations within the dataset, which can be regarded as false negative cases. Other 49 cases are true-negative cases and occurred due to the presence of \textbf{DN}. Regarding the limitations within the dataset, we classify them into three distinct types: \textbf{Equivalent Answer Example}, \textbf{Alternative Answer}, and \textbf{Temporality}.

\subsubsection{Dataset Issue}
Dataset issues exemplify situations where the predicted answers are accurate, yet the evaluation is insufficient. Prediction within the context is highlighted in {\color{red}red}, while the supporting information is marked in {\color{blue}blue}.
\begin{itemize}
  \item \textbf{Equivalent Answer Example} \\
    There are 34 cases where predictions have different formats or are supersets/subsets of gold answers. 
        \begin{quote}
        \rule{\linewidth}{2pt}
        \textbf{Query}: when did the king kong ride burn down \\
        \textbf{EM pattern}: 01000000000000000000 \\
        \textbf{Gold Answers}: ['2008'] \\
        \rule{\linewidth}{1pt}
        \textbf{DN index}: 2 \\
        \textbf{Prediction} : June 1, 2008 \\
        \textbf{Predicted answer in DN context}: Yes \\
        \textbf{DN context} \\
        \{'id': '4215890', 'title': 'King Kong', 'text': 'and Universal Orlando Resort in Orlando, Florida. The first King Kong attraction was called King Kong Encounter and was a part of the Studio Tour at Universal Studios Hollywood. Based upon the 1976 film "King Kong", the tour took the guests in the world of 1976 New York City, where Kong was seen wreaking havoc on the city. It was opened on June 14, 1986 and {\color{blue}was destroyed on {\color{red}June 1, 2008} in a major fire}. Universal opened a replacement 3D King Kong ride called "" that opened on July 1, 2010, based upon Peter Jackson\'s 2005 film "King Kong".'\} \\
        \rule{\linewidth}{2pt}
        \end{quote}
  \item \textbf{Alternative Answer Example} \\
    There are 13 cases in which the predictions can serve as alternative answers for the given query
        \begin{quote}
        \rule{\linewidth}{2pt}
        \textbf{Query}: who introduced the system of civil services in india \\
        \textbf{EM pattern}: 00011111111100000000 \\
        \textbf{Gold Answers}: ['Charles Cornwallis'] \\
        \rule{\linewidth}{1pt}
        \textbf{DN index}: 12 \\
        \textbf{Prediction} : Warren Hastings \\
        \textbf{Predicted answer in DN context}: No \\
        \rule{\linewidth}{1pt}
        \textbf{Context index including prediction}: 3 \\
        \textbf{Context} \\
        \{'id': '14394957', 'title': 'Civil Services of India', 'text': "administer them. The civil service system in India is rank-based and does not follow the tenets of the position-based civil services. In 2015, the Government of India approved the formation of Indian Skill Development Service. Further, in 2016, the Government of India approved the formation of Indian Enterprise Development Service. {\color{red}Warren Hastings} {\color{blue}laid the foundation of civil service} and Charles Cornwallis reformed, modernised and rationalised it. Hence, Charles Cornwallis is known as the 'Father of Civil Service in India'. He introduced Covenanted Civil Services (Higher Civil Services) and Uncovenanted Civil Services (Lower Civil Services). The present civil services of India"\} \\
        \rule{\linewidth}{2pt}
        \end{quote}
  \item \textbf{Temporality Example} \\
    There are 4 cases where queries ask for the up-to-date context but generate the answer on the outdated context
        \begin{quote}
        \rule{\linewidth}{2pt}
        \textbf{Query}: who plays cat in beauty and the beast \\
        \textbf{EM pattern}: 01111111111111111110 \\
        \textbf{Gold Answers}: ['Kristin Kreuk'] \\
        \rule{\linewidth}{1pt}
        \textbf{DN index}: 19 \\
        \textbf{Prediction} : Linda Carroll Hamilton \\
        \textbf{Predicted answer in DN context}: No \\
        \rule{\linewidth}{1pt}
        \textbf{Context index including prediction}: 6 \\
        \textbf{Context} \\
        \{'id': '728625', 'title': 'Linda Hamilton', 'text': 'Linda Hamilton {\color{red}Linda Carroll Hamilton} (born September 26, 1956) is an American actress best known for her portrayal of Sarah Connor in "The Terminator" film series and {\color{blue}Catherine Chandler in the television series "Beauty and the Beast" (1987-1990)}, for which she was nominated for two Golden Globe Awards and an Emmy Award. She also starred as Vicky in the horror film "Children of the Corn". Hamilton had a recurring role as Mary Elizabeth Bartowski on NBC\'s "Chuck". Hamilton was born in Salisbury, Maryland. Hamilton\'s father, Carroll Stanford Hamilton, a physician, died when she was five, and her mother later married'\} \\
        \rule{\linewidth}{2pt}
        \end{quote}
\end{itemize}

\subsubsection{Definite Negative Cases}
The presence of \textbf{DN} contexts destabilize the reader, leading to the conversion of correct gold answer.  
The prediction present within the context is highlighted in {\color{red}red}.
\begin{itemize}
  \item \textbf{Predictions in DN context} \\
    In 43 cases, the predictions (transitioned from gold answers) are found within the DF passage.
    \begin{quote}
    \rule{\linewidth}{2pt}
    \textbf{Query}: real name of gwen stacy in amazing spiderman \\
    \textbf{EM pattern}: 11110011111111111111 \\
    \textbf{Gold Answers}: ['Emma Stone'] \\
    \rule{\linewidth}{1pt}
    \textbf{DN index}: 4 \\
    \textbf{Prediction} : Mary Jane Watson \\
    \textbf{Predicted answer in DN context}: Yes \\
    \textbf{DN context} \\
    \{'id': '1283490', 'title': 'Gwen Stacy', 'text': 'relationship with chemical weapon developer Norman Osborn. {\color{red}Mary Jane Watson}, a popular actress in this reality, played Gwen Stacy in the film adaptation of Spider-Man's life story. Gwen and her father read textual accounts of their deaths in the main universe, though they believe this simply to be the morbid imaginings of Peter Parker, who is suffering from mental health issues. Gwen Stacy first appeared in "Marvel Adventures Spider-Man" \#53 as a new student of Midtown High. She had transferred from her previous school after the Torino Gang, a powerful New York mob, began harassing her in an attempt to ' \} \\
    \rule{\linewidth}{2pt}
    \end{quote}
  \item \textbf{Predictions in previous context} \\
  There are 4 cases where the prediction appears in the previous contexts.
    \begin{quote}
    \rule{\linewidth}{2pt}
    \textbf{Query}: who will get relegated from the premier league 2016/17 \\
    \textbf{EM pattern}: 00000100011111101111 \\
    \textbf{Gold Answers}: ['Middlesbrough', 'Sunderland', 'Hull City'] \\
    \rule{\linewidth}{1pt}
    \textbf{DN index}: 15 \\
    \textbf{Prediction} : Norwich City \\
    \textbf{Predicted answer in DN context}: No \\
    \rule{\linewidth}{1pt}
    \textbf{Context index including prediction}: 7 \\
    \textbf{Context} \\
    \{'id': '19245453', 'title': '2016–17 Premier League', 'text': "the league – the top seventeen teams from the previous season, as well as the three teams promoted from the Championship. The promoted teams were Burnley, Middlesbrough and play-off winners Hull City, who replaced Aston Villa, {\color{red}Norwich City} and Newcastle United. West Ham United played for the first time at the London Stadium, formerly known as the Olympic Stadium. Although having a capacity of 60,010, for the first Premier League game this was limited to 57,000 due to safety fears following persistent standing by fans at West Ham's Europa League game played in early August. Stoke City announced that from"\}\\
    \rule{\linewidth}{2pt}
    \end{quote}
  \item \textbf{Predictions not in contexts} \\
    There are 2 cases where the prediction does not exist in the candidate passages.
    \begin{quote}
    \rule{\linewidth}{2pt}
    \textbf{Query}: how many episodes does riverdale season one have \\
    \textbf{EM pattern}: 11110000111111111111 \\
    \textbf{Gold Answers}: ['13'] \\
    \rule{\linewidth}{1pt}
    \textbf{DN index}: 4 \\
    \textbf{Prediction} : 21 \\
    \textbf{Predicted answer in DN context}: No \\
    \rule{\linewidth}{1pt}
    \textbf{Context index including prediction}: None \\
    \rule{\linewidth}{2pt}
    \end{quote}

\end{itemize}

\subsection{Passage Type Classification}
\label{appen:psgc}

\subsubsection{Mutli-classification}
\label{appen:multi-class}
\begin{table}[h!]
\small
\centering
\begin{tabular}{clcccccc}
\hline
Model               & \multicolumn{1}{l|}{Metric}             & DP  & DN  & SP   & SN     & IZ   \\ \hline
                      & \multicolumn{1}{l|}{precision}        & 0   & 0   & 0.54 & 0.09   & 0.5       \\
\multirow{-2}{*}{FiD-Encoder}  & \multicolumn{1}{l|}{recall}  & 0   & 0   & 0.72 & 0.04   & 0.37       \\ \hline
                      & \multicolumn{1}{l|}{precision}        & 0   & 0   & 0.55 & 0.1    & 0.5   \\
\multirow{-2}{*}{T5-Encoder} & \multicolumn{1}{l|}{recall}    & 0   & 0   & 0.72 & 0.05   & 0.37      \\ \hline
                      & \multicolumn{1}{l|}{precision}        & 0   & 0   & 0.5  & 0      & 0   \\
\multirow{-2}{*}{RoBERTa} & \multicolumn{1}{l|}{recall}       & 0   & 0   & 1    & 0      & 0    \\ \hline \hline
\multicolumn{2}{c|}{Total Instances}                          & 1308 & 396   & 86936     & 15438   & 71122    \\ \hline

\end{tabular}
\caption{The table shows the results on 175,200 evaluations instances, which are 20\% of original DPR-retrieved NQ devset. All models struggle with identifying \textbf{DN} and \textbf{DP}, favoring predictions of \textbf{SP}, which constitutes 50\% of instances. Notably, all three models can't differentiate context types properly. Among 175,200 total instances, T5 predicts 114,606 (65.4\%) to SP, FiD labels 115,522 (65.9\%) as SP, and RoBERTa classifies all as SP. This highlights multi-classification's challenge in capturing passage types and inability to identify \textbf{DN} and \textbf{DP}.
}
\label{appen4:multi}
\end{table}

\subsubsection{End-to-End Results}
\label{appen:e2e}
\begin{table}[h!]
\small
\centering
\begin{tabular}{clcccccc}
\hline
& \multicolumn{1}{l|}{Model}  & Probe 1 & Probe 2 & Probe 3 & Probe 4 & Probe 5 & Probe 6 \\ \hline
& \multicolumn{1}{l|}{FiD-Encoder}  & 48.59   & 37.51 & 53.77 & 42.74    & 54.71   & 43.66  \\
& \multicolumn{1}{l|}{T5-Encoder} & 48.73   & 37.45   & 53.46 & 42.3   & 54.24   & 43.19   \\
& \multicolumn{1}{l|}{RoBERTa} & 54.43   & 35.84   & 54.43 & 35.84   & 54.43   & 35.84   \\ \hline
\end{tabular}
\caption{End-to-End result of multi-classifcation on NQ test set. Among all classification models, RoBERTa achieves the highest EM scores across Probe1, Probe3, and Probe5. This is due to the fact that RoBERTa inferences are all \textbf{SP}s, resulting in all contexts being used as inputs for the reader. Consequently, these scores should match EM@100(54.43), which uses all contexts for inference. All Probe results exhibit either lower or equal performance compared to EM@100, notably falling significantly below Probe 3's EM score(65.12). This underscores the classification models' struggle in identifying EM patterns and passage types, reflecting the challenge in distinguishing \textbf{DN} and \textbf{DP} within the provided retrieved context.
}
\label{appen5:end2end}
\end{table}

\clearpage

\subsection{Semantically Equivalent Answers}
\label{appen:sem_em}
To calculate the SeEM and SeAcE, candidate answers are collected from the incremental answers. Then, we prompt \textbf{gpt-3.5-turbo-0613} and \textbf{Claude2} by iterating the candidate answer list over the gold answer list. Suppose we have an instance like this:

\begin{small}
\begin{verbatim}
    query = "who sings does he love me with reba"
    gold_ans_lst = ['Linda Davis']
    cand_lst = ['Linda Kaye Davis', 'Linda Davis']
\end{verbatim}
\end{small}

In this case, EM fails to capture "Linda Kaye Davis" as the correct answer because of  "Kaye" in the middle.
We perform zero-shot-prompting by iterating over \textbf{gold\_ans\_lst} and \textbf{cand\_lst}. Here is an example:

\begin{small}
\begin{verbatim}
    Question: who sings does he love me with reba
    Answer: Linda Davis
    Candidate: Linda Kaye Davis
    Is candidate correct?
\end{verbatim}
\end{small}

Although each API generates different response formats to the given prompt, we only consider cases where the response starts with explicit tokens (Yes/No); otherwise, we discard the candidate.

\begin{small}
\begin{verbatim}
    response: Yes, the candidate is correct. 
    Linda Kaye Davis is the singer who performed the duet "Does He Love You" with Reba McEntire.
\end{verbatim}
\end{small}

Finally, we construct an adjusted gold answer list based on the response and calculate the SeAm and SeAcEm.

\subsection{Attention Score of top-20 passges}
\label{append:attention-score}
\begin{figure}[!h]
  \centering
  \includegraphics[width=0.75\linewidth]{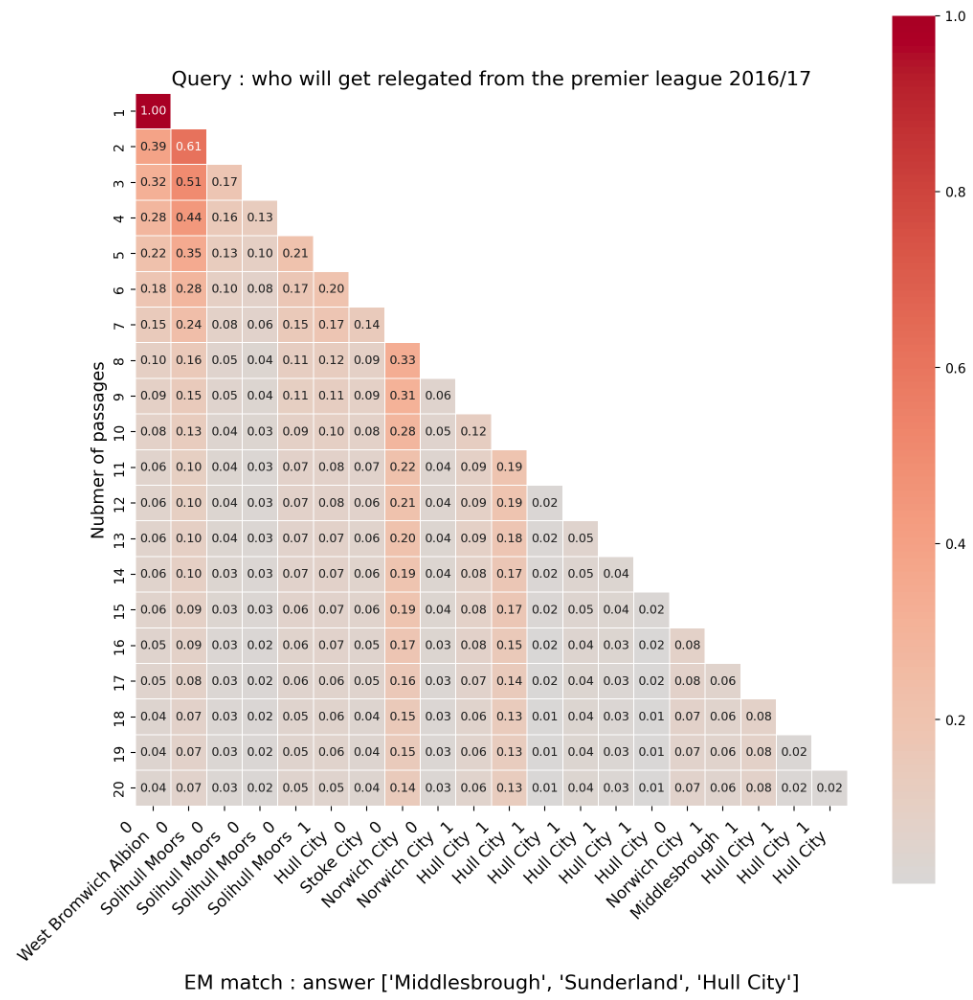}
  \caption{The attention scores of the top 20 retrieved passages are depicted in this example. It can be observed that the 7th and 16th passages are \textbf{DN}. Interestingly, the passage with the highest attention score(8th) is identified as \textbf{SN} }
  \label{fig:attention-score-large}
\end{figure}

\end{document}